\documentclass[journal]{IEEEtran}
\usepackage{graphicx}
\usepackage{booktabs}
\usepackage{subcaption}
\usepackage{amssymb}
\usepackage{multirow}
\usepackage{dsfont }
\usepackage{xspace}
\usepackage{amsmath}

\usepackage[pagebackref=true,breaklinks=true,letterpaper=true,colorlinks,bookmarks=false]{hyperref}
\usepackage{cite}

\ifCLASSINFOpdf
\else
\fi
\begin{document}

\title{Few-shot Segmentation with Optimal Transport Matching and Message Flow}

\author{Weide~Liu,
        Chi~Zhang,
        Henghui~Ding,
        Tzu-Yi~Hung and
        Guosheng~Lin

\thanks{W. Liu is with School of Computer Science and Engineering, Nanyang Technological University (NTU), Singapore 639798 and with Institute for Infocomm Research, A*STAR, Singapore 138632 (e-mail: weide001@e.ntu.edu.sg).}

\thanks{C.~Zhang is with School of Computer Science and Engineering, Nanyang Technological University (NTU), Singapore 639798 (e-mail: chi007@e.ntu.edu.sg).}

\thanks{H.~Ding is with School of Electrical and Electronic Engineering, Nanyang Technological University (NTU), Singapore 639798 (e-mail: ding0093@e.ntu.edu.sg).}

\thanks{T.~Hung is with Delta Research Center, Singapore (e-mail:  tzuyi.hung@deltaww.com).}

\thanks{G.~Lin is with School of Computer Science and Engineering, Nanyang Technological University (NTU), Singapore 639798 (e-mail: gslin@ntu.edu.sg).}

\thanks{Corresponding author: Guosheng Lin.}

}

\markboth{IEEE TRANSACTIONS ON MULTIMEDIA}%
{Shell \MakeLowercase{\textit{et al.}}: Bare Demo of IEEEtran.cls for IEEE Journals}

\maketitle

\begin{abstract}
We tackle the challenging task of few-shot segmentation in this work. It is essential for few-shot semantic segmentation to fully utilize the support information. Previous methods typically adopt masked average pooling over the support feature to extract the support clues as a global vector, usually dominated by the salient part and lost certain essential clues. In this work, we argue that every support pixel's information is desired to be transferred to all query pixels and propose a Correspondence Matching Network (CMNet) with an Optimal Transport Matching module to mine out the correspondence between the query and support images. Besides, it is critical to fully utilize both local and global information from the annotated support images.
To this end, we propose a Message Flow module to propagate the message along the inner-flow inside the same image and cross-flow between support and query images, which greatly helps enhance the local feature representations. 
Experiments on PASCAL VOC 2012, MS COCO, and FSS-1000 datasets show that our network achieves new state-of-the-art few-shot segmentation performance.
\end{abstract}

\begin{IEEEkeywords}
Few-shot Learning, Segmentation, Correspondence Matching Network, CMNet, Optimal Transport Matching, Message Flow
\end{IEEEkeywords}

\IEEEpeerreviewmaketitle

\section{Introduction}
\IEEEPARstart{S}{emantic} segmentation is a fundamental task to assign every pixel with a label. With the rapid development of deep learning, fully supervised semantic segmentation has significantly improved in recent years. One of the intrinsic limitations of the fully supervised segmentation is that it requires large amounts of annotated images to train the model. Another limitation is that the performance will plummet when predicting unseen classes. 

\begin{figure}[t]
\centering
    \includegraphics[width=1\linewidth]{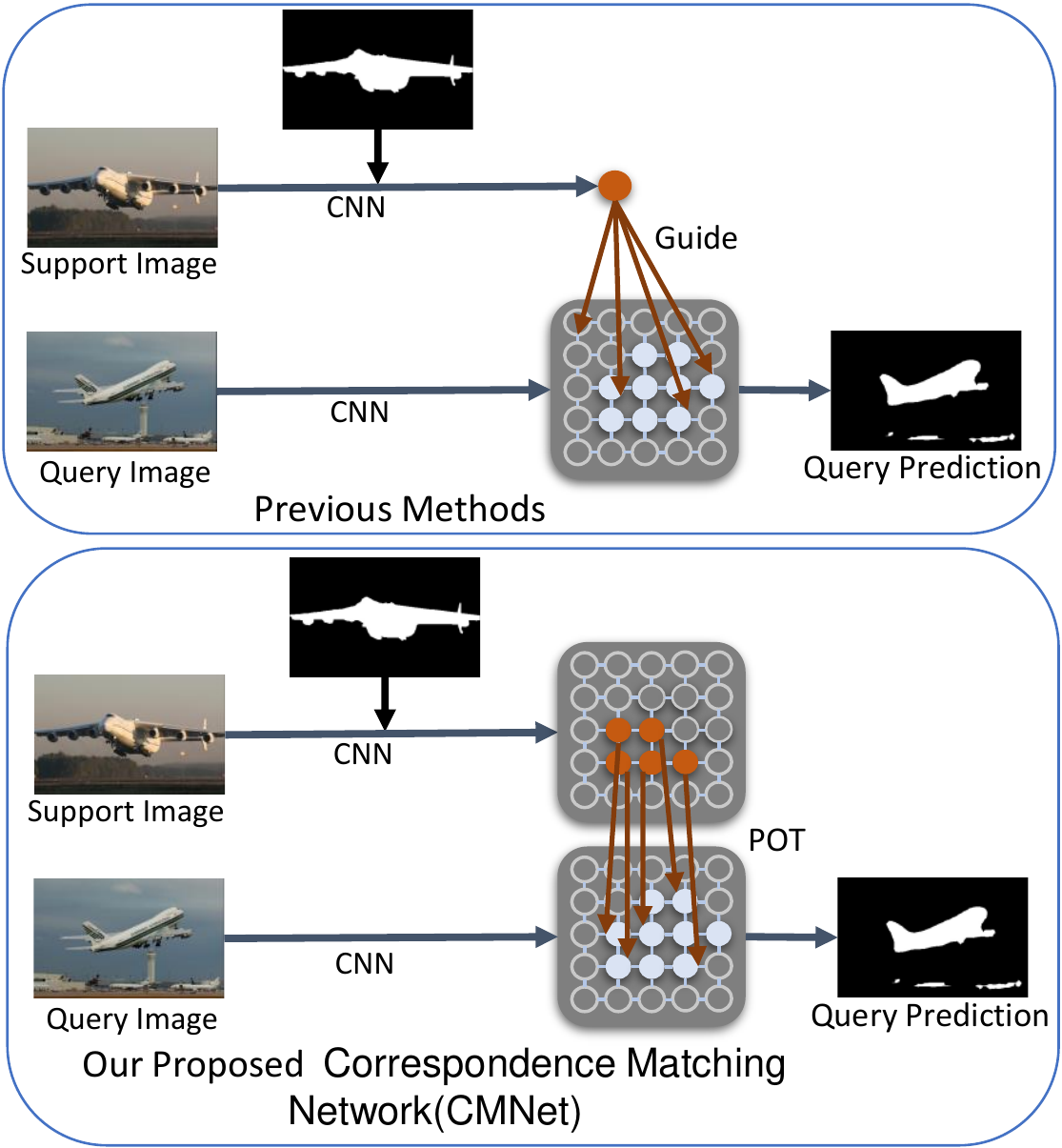}
    \caption{Comparison between the pipeline of our proposed network with the previous methods of few-shot segmentation. Previous works (upper part) average the support features to a global vector to guide the query image’s predictions. In contrast, our proposed CMNet (lower part) guides each query feature with specific clues by applying a partial optimal transport matching (POT) between query and support features.}
    \label{Fig:motivation}
\end{figure}

\begin{figure*}[t]
\centering
    \includegraphics[width=1\linewidth]{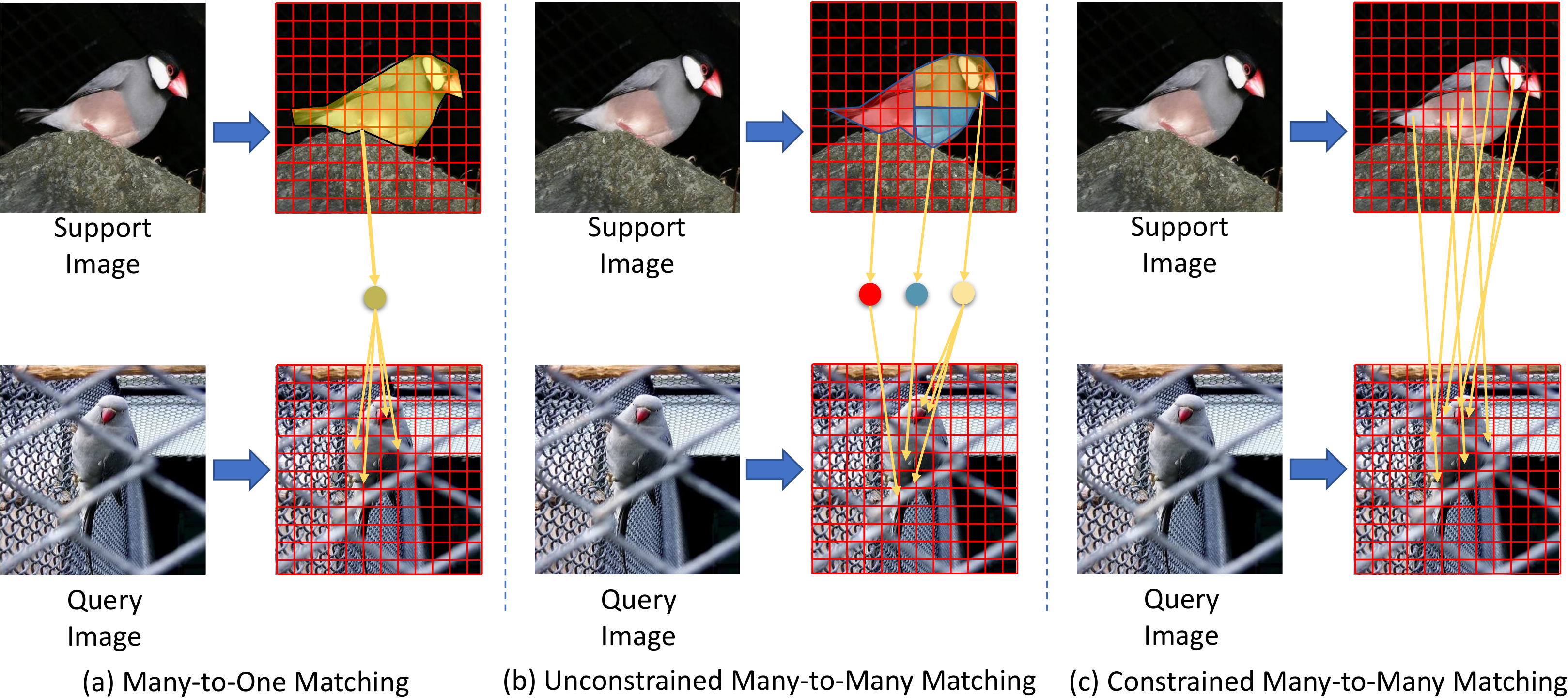}
    \caption{The comparison of diffident matching methods. Many-to-one matching methods (Figure a) average the support features to a global vector to guide the query image’s predictions, which results in an uncertain connection. The unconstrained many-to-many methods (Figure b) make the network attend to the most discriminative parts, resulting in many unmatched query features. In contrast, our proposed constrained many-to-many matching method (optimal transport matching) (Figure c) encourages the network to mine more related correspondences.}
    \label{Fig:diff-matching}
\end{figure*}

To address the above issues, few-shot segmentation is proposed to learn a network that can predict novel classes with only a few annotated data. Few-shot segmentation divides the data into support sets and query sets. The goal is to predict a binary mask in query images. The categories provided by the annotated support sets should be predicted as foreground. Two-branch network architecture has been widely used~\cite{zhang2019canet,crnet,wang2019panet} to convey the information between the query and support images.
However, previous methods typically adapt masked average pooling over the support feature to extract the support clues as a global vector, usually dominated by the salient part and missing certain key clues. Besides, the pixel relations between the image pair are unclear with merely a global vector. As shown in Fig.~\ref{Fig:diff-matching}, previous methods~\cite{zhang2019canet,crnet} simplify the many-to-many problem to the many-to-one problem by averaging the support features to a global vector to guide the query image's predictions. In this way, every position for the query image will have the same clue. However, one obvious limitation is that it breaks the spatial structures and only captures the most dominant features of support images, resulting in limited parts of the support images being used to match the query images. 
We argue that every support pixel's information is desired to be transferred to all query pixels, resulting in a many-to-many message problem \cite{zhangchi2}.

PGNet~\cite{zhangchi2} is a many-to-many work that proposes using pyramid graph connections to transport information between the query and support sets with a structured representation. However, there are several limitations to such unconstrained many-to-many matching methods: 1) \textbf{many-to-one matching.} A feature point from the query image may have more than one correspondence point in the support image, which results in that only the most discriminative parts of the query features connect to the support features~\cite{danet}.
2) \textbf{Inappropriate matching.} The dominant parts in the query image connect too many points in the support image(including the dominant and the normal parts), resulting in inappropriate connections. For example, as shown in Figure~\ref{Fig: pg_ours}(Figure b), some background of the query image has been highlighted as the correspondences. 
3) \textbf{Unmatched parts.} Many query features may be unmatched due to the \textit{many-to-one matching} problem. For example, as shown in Fig.~\ref{Fig:diff-matching}, most features from the support image match the head of the bird in the query image; consequently, most regions of the query image remain unmatched. 

To mine out more correspondences between the query and support images and address the limitations of the previous graph network, we propose a constrained many-to-many matching method, partial optimal transport matching, to maximize the total correspondences between the query and support images. We fix the maximum matching flow at a value according to the object size extracted from the ground truth of the support image. This will stimulate the query feature to activate the object parts whose features are less discriminative but necessary to be matched. For example,
we need to match the features between two birds in Fig.~\ref{Fig:diff-matching}.
In the previous works, the most discriminative part \textit{head} will be matched, but the remaining parts are not able to be matched.
In contrast, our proposed partial optimal transport matching module could activate the less discriminative parts by setting a maximum matching flow based on the object size.

Furthermore, previous approaches such as~\cite{crnet,zhang2019canet,wang2019panet} focus more on global features while relatively overlooking the information of the local feature representations. We spotlight the importance of the local representations for the few shot segmentation tasks. To enhance the local feature representations, we develop a message flow module to propagate messages within one image (inner-flow) and between various images (cross-flow). We model the query and support features using graphs and propagate the information among them. The nodes are associated with local features, and the edges are associated with the similarity information between nodes. 
Our optimal transport matching module also benefits from the enhanced local feature representations to generate the correspondences.

We handle the domain gap between different datasets for the few-shot segmentation task by learning the network parameters from a fully supervised segmentation task with the base data. Previous works~\cite{zhang2019canet,crnet,zhangchi2,danet,pfenet,few-shot-1,few-shot-2,few-host-3,few-host-4} use frozen backbones pretrained on ImageNet~\cite{imagenet}. However, there are domain gaps between ImageNet and the target datasets. We find that pre-training on the base class data will assist the model in recognizing the new objects and mitigate the data domain gap issue.

Our contributions are summarized as follows:
\begin{itemize}
    \item We propose a partial optimal transport matching module, which is a constrained many-to-many matching method, for the few shot segmentation tasks. The query and support images are allowed to match more related areas as correspondences. 
    \item We develop a message flow module to propagate the message between images along the inner-flow and cross-flow to enhance the local feature representations. 
    \item Experiments on PASCAL VOC 2012, MS COCO, and FSS-1000 dataset show that our method outperforms the baseline and achieves new state-of-the-art results.
\end{itemize}

\section{Related Work}
\subsection{Semantic Segmentation}
Semantic segmentation is a fundamental computer vision task that aims to assign each pixel with a class label. Currently, the most popular architecture is fully convolutional networks (FCN)~\cite{long2015fully} which replaces the final fully-connected layer of conventional classification network with a $1\times1$ convolutional layer and has been widely used in current state-of-the-art methods, including ours.
Encoder-decoder~\cite{refine-tpami,crnet,chen2018deeplab,tmm-shi2018hierarchical,tmm-zhang2019decoupled,tmm-zhou2020sal,liu2020guided,liu2020weakly,zhang2020splitting,liu2021few_global_local,liu2021cross,liu2022long,hou2022distilling,hou2020attention,hou2020attention,hou2020object} is a popular structure used to reconstruct the high-resolution prediction map for semantic segmentation prediction. The encoder gradually downsamples the feature maps to obtain a large field-of-view, and the decoder recovers the fine-grained information. Dilated convolution~\cite{chen2018deeplab} also aims to increase the field-of-view by increasing the convolution dilation without increasing the parameters. The cross-image relation~\cite{wang2,wang3} has been explored in segmentation tasks, which is similar to the few-shot segmentation.
In our network, we also follow the encoder-decoder structure with a dilated convolution to explore the cross-image relation for our semantic segmentation prediction.

\subsection{Few-shot Segmentation}
Metric learning has been widely used in few-shot segmentation tasks. Such as CANet~\cite{zhang2019canet} uses the global feature from the support image to make a dense comparison to the query image. OSLSM~\cite{shaban2017one} introduces a two-branch network consisting of support and query branches for few-shot segmentation, the support branch designed to guide the query set. CRNet~\cite{crnet} proposes that the query branch and the support branch can guide each other with a cross-reference mechanism. PL~\cite{Dong2018FewShotSS} employs the prototypical network for few-shot segmentation as metric learning. SG-One obtains a prototype vector from the support image to guide query prediction with a masked-average pooling. PANet~\cite{wang2019panet} introduces a prototype alignment regularization between support and query images and guides each other. 

Most of those metric learning for few-shot segmentation generate the prototype vector with global average pooling to guide the query image. However, such a strategy has two limitations: spatial information has been disregarded; another is that the objects' semantic information is underutilized. To fully use spatial information, we adopt the message flow module to propagate the local information between images and enhance local feature representations. 

\subsection{Optimal transport}
Optimal transport provides a way to formulate the problem of transferring one distribution to another as linear programming, which can be solved by Sinkhorn algorithm~\cite{sinkhorn-algrithm}. Optimal transport has been widely used in many various computer vision tasks. Courty~\textit{et al.}~\cite{courty2016optimal} learns a transportation plan to step over the domain gap between the source domain to the target domain. Su~\textit{et at.}~\cite{su2015optimal} matches the 3D shape with the optimal transport. Zhao~\textit{et al.}~\cite{zhao2008differential} employs a differential-optimal transport to tackle the visual tracking problem. Zhou~\textit{et al.}~\cite{wang-part} proposed to solve the partial matching problem with the Projected Gradient Descent.
In this paper, we use a partial optimal transport module to obtain the semantic correspondences between the query and support images.

\subsection{Self-supervised Learning}
The self-supervised learning methods have been widely used in computer vision tasks. For example, the self-supervised contrastive learning methods~\cite{moco,Bootstrap} have been proposed to learn a feature extractor from unlabeled images as pre-training for other downstream tasks. The contrastive learning methods aim to pull close the representations of different views of the same image and push away the representations of different images. The LIIR~\cite{wang-self} proposes self-supervised correspondence learning to obtain better feature representations from unlabeled videos. 
Our method utilizes correspondence matching for the few shot segmentation tasks.
Few-shot segmentation task aims to predict binary masks to segment out the foreground objects in the testing image set with only a few annotated images. The categories in the training image set have no overlap with the testing image set. The annotated images are treated as support sets for the training and testing interface and provide the foreground categories to guide the segmentation of the unlabeled query set. 
\section{Methods}
 \begin{figure*}[t]
 \centering
    \includegraphics[width=1\linewidth]{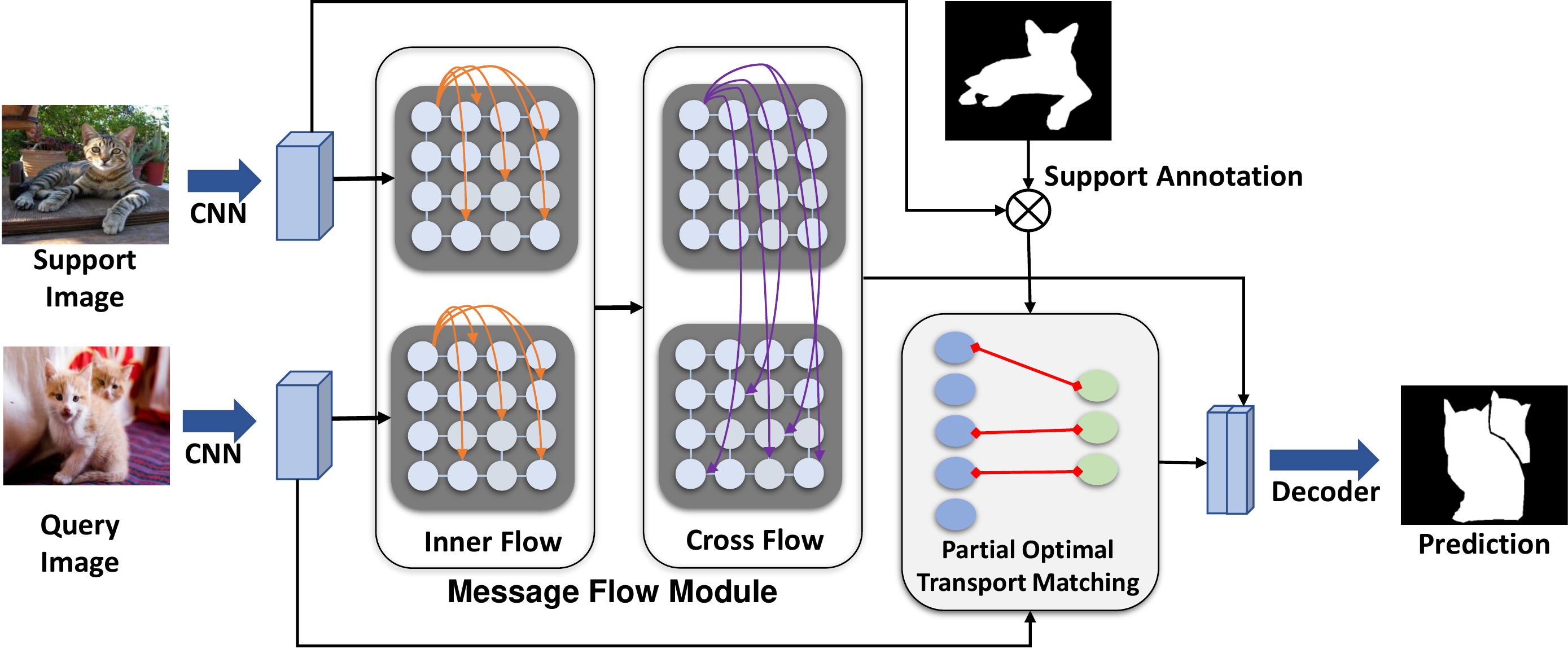}
    \caption{Our network architecture. Our network mainly consists of a message flow module and a partial optimal matching module. We encode every support and query image pair into graphs with a shared-parameter CNN. Then we propagate the message between the graph nodes using the inner-flow (inside one image) and cross-flow (across images) to enhance local features. Combining the optimal matching correspondences generated by our optimal matching module, we generate the query predictions with our decoder.}
    \label{archtecture}
\end{figure*}
The critical technique for few-shot semantic segmentation is the way to propagate the local and global information between the query and support images. 
Previous methods merely average the support feature to a global vector, serving as a general clue to guide every position of the query image. However, the global vector lacks the information of spatial structures and only captures the most dominant information of the support features, resulting in the limited parts of the support images being used to match the query images.  
The unconstrained many-to-many matching methods attempt to solve the problem by establishing a weighted connection between each query and support features. 
However, due to the many-to-one problem, most of the features from the support image will match the most discriminative parts in the query image, which causes the remaining query features to be unmatched.

This section introduces a constrained many-to-many matching method to maximize the total correspondences between the query and support images to overcome the aforementioned flaws. Then we will introduce a message flow module to propagate messages within one image (inner-flow) and between different images (cross-flow) to enhance the local feature representations. Our optimal transport module also benefits from the enhanced local feature representations to generate the correspondences.
Finally, we introduce how to overcome the domain gap between different datasets by learning the network parameters from a fully supervised segmentation task with the base data. The overall network architecture is shown in Fig.~\ref{archtecture}.

\subsection{Partial Optimal Transport Module}
\textbf{Optimal Transport.}
Optimal transport aims to minimize the distance to transfer the source image distribution to the target distribution, which provides a way to generate the correspondence between two distributions. Specially, we treat the source distribution as the suppliers and the target as demanders. Suppose there are a set of suppliers $\mathcal{S} = \{s_i| i = 1,2,...m\}$ required to transport goods to the demanders $\mathcal{D} = \{d_j|j = 1,2,...k\}$, and the cost per unit transport from each supplier to demander denotes $c_{ij}$. The goal of optimal transport is to cost as little as possible to transport as many goods as possible:
\begin{equation}
    \underset{x_{ij}}{\text{minimize}}\sum\nolimits_{i=1}^{m}\sum\nolimits_{j=1}^{k}c_{ij}x_{ij}. 
\end{equation}
where $x_{ij}$ denotes the number of goods from $s_i$ to $d_j$. Note that the total goods from the supplier should be equal to the demanders' demands, and the goal can be achieved by solving the linear programming problem. 

\textbf{Partial Optimal Transport.}
The total amount of goods provided by the supplier is not always equal to the goods demanders required. To address this problem, partial optimal transport allows matching two asymmetric distributions with some absolute amount quantity, and we call this \textit{absolute partial optimal transport}. In an absolute partial matching problem, the amount of weight-matched is defined as $M$, and the total transport flow will be equal to $M$:
\begin{equation}
\begin{aligned}
&x_{ij}\geqslant 0, i=1,...,m,j=1,...,k\\
&\sum\nolimits_{j=1}^{k}x_{ij} <= w_i,  ~ ~ i=1,...m \\
&\sum\nolimits_{i=1}^{m}x_{ij} <= u_j, ~ ~  j=1,...k \\
&\sum\nolimits_{i=1}^{m}\sum\nolimits_{j=1}^{k}x_{ij}=M
\end{aligned}
\label{emd_ori}
\end{equation}
where $w_i$ denotes the number of goods the $i_{th}$ supplier can provide and the $u_j$ denotes the goods for $j_{th}$ demander required. To achieve a partial matching goal, we treat this problem as a balanced transportation problem. Suppose total $k$ demanders require $w_d$ goods, while all the $m$ suppliers can provide $w_s$ goods(suppose $w_d >= w_s$), and a fixed amount of transport weight $M <= w_s$ is given. To balance the goods amount between supplier and demanders, we set a balanced flow $M(\gamma) = w_d + w_s - M $. Then we set a $m+1$ supplier as a dummy supplier to provide the supply a unbalanced weight $M(\gamma) - w_s = w_d -M$ and a dummy demander $k+1$ to requires the unbalance demand weight $M(\gamma) - w_d = w_s - M$. Combined with the dummy supplier and demander, the balanced transportation problem can be solved as an optimal transport problem. We transfer the partial matching problem to a balanced transportation problem in the following way:
\begin{equation}
\begin{aligned}
&w_s = \sum\nolimits_{i=1}^{m}s_i, ~ ~ i=1,...,m\\
&w_d = \sum\nolimits_{j=1}^{k}d_j, ~ ~ j=1,...,k\\
&M(\gamma) = w_d + w_s - M \\
&w_{m+1} = w_d -M \\
&u_{k+1} = w_s -M \\
&c_{ij} >= 0 \\
&c_{m+1,j} = 0, ~ ~  j=1,...,k\\
&c_{i,k+1} = 0, ~ ~  i=1,...,m\\
&c_{m+1.n+1} = \infty .
\label{eq: part-emd}
\end{aligned}
\end{equation}
We set the dummy supplier's cost to the dummy demander to $\infty$ to prevent matching the dummy nodes. Then we set the cost from dummy supplier $s_{m+1}$ to the normal demanders at zero ($c_{m+1,j} =0 $), and the dummy demander $d_{k+1}$ also no cost to match the remaining suppliers($c_{i,k+1} =0 $). Therefore, the normal suppliers will firstly provide $w_s -M$ goods to the dummy demander and remain $M$ goods to match the normal demanders. On the other hand, the normal demanders will receive $wd - M$ goods from the dummy supplier as prior, and remain $M$ goods will be provided by the normal suppliers. Therefore, only transport weight $M$ will be matched with a nonzero cost between the normal suppliers and demanders. In this way, the partial optimal transport problem has been formulated as a balanced transportation problem to find the optimal flow from suppliers $s_i, i = 1,2..m$ to demanders $s_j, j = 1,2..k$, which can be solved the same as the full optimal transport problem. 

\textbf{Optimal Transport for Few-shot Segmentation.}
Metric learning methods are widely used to tackle the few-shot segmentation task. CRNet and CANet generate a global support vector~\cite{zhang2019canet,crnet} to find an optimal vector to \textbf{match} all the elements from the query set. However, those methods ignore the local discriminative representations from the support images. PANet~\cite{wang2019panet} generates a cosine similarity map with local information as the final prediction, while PFENet~\cite{pfenet} generates a prior mask by obtaining the pixel-wise correspondence from the cosine similarity maps. 

Different from the previous works, our method not only takes the pixel-wise correspondence value from the cosine similarity maps as an intermediate representation; inspired by previous methods~\cite{sarlin2020superglue,zhang2020deepemd} we also find the correspondences between the query and support images by solving a \textbf{many-to-many} matching problem. 

In particular, we encode each image into features and model as a graph. Each local representation denotes a node. The nodes from support images are set to suppliers $s_i$ and set the query nodes as demanders $d_j$. The cost between suppliers and demanders depends on their cosine similarity: 
\begin{equation}
c_{ij}=1-\frac{{\mathbf{s}_i}^T  \mathbf{d}_j}{\lVert \mathbf{s}_i\rVert \lVert \mathbf{d}_j\rVert}.
\label{eq: cost}
\end{equation}
In this way, we formulate the few-shot segmentation problem into a partial optimal matching problem as described above. By using the Sinkhorn algorithm~\cite{sinkhorn-algrithm}, we obtain an optimal flow from query to support image. The flow weight from each query node to the support node represents the correspondence between the nodes. By multiplying the binary support annotation to the support nodes, we filter out the irrelevant correspondences from the query image to the support image. The foreground probability maps will be generated with the flitted correspondences, while the correspondences' weight can be considered the foreground probability. Finally, we fuse the foreground probability map as a coarse object region to guide the feature maps to segment the final query masks. 

\begin{figure*}[t]
\centering
    \includegraphics[width=1\linewidth]{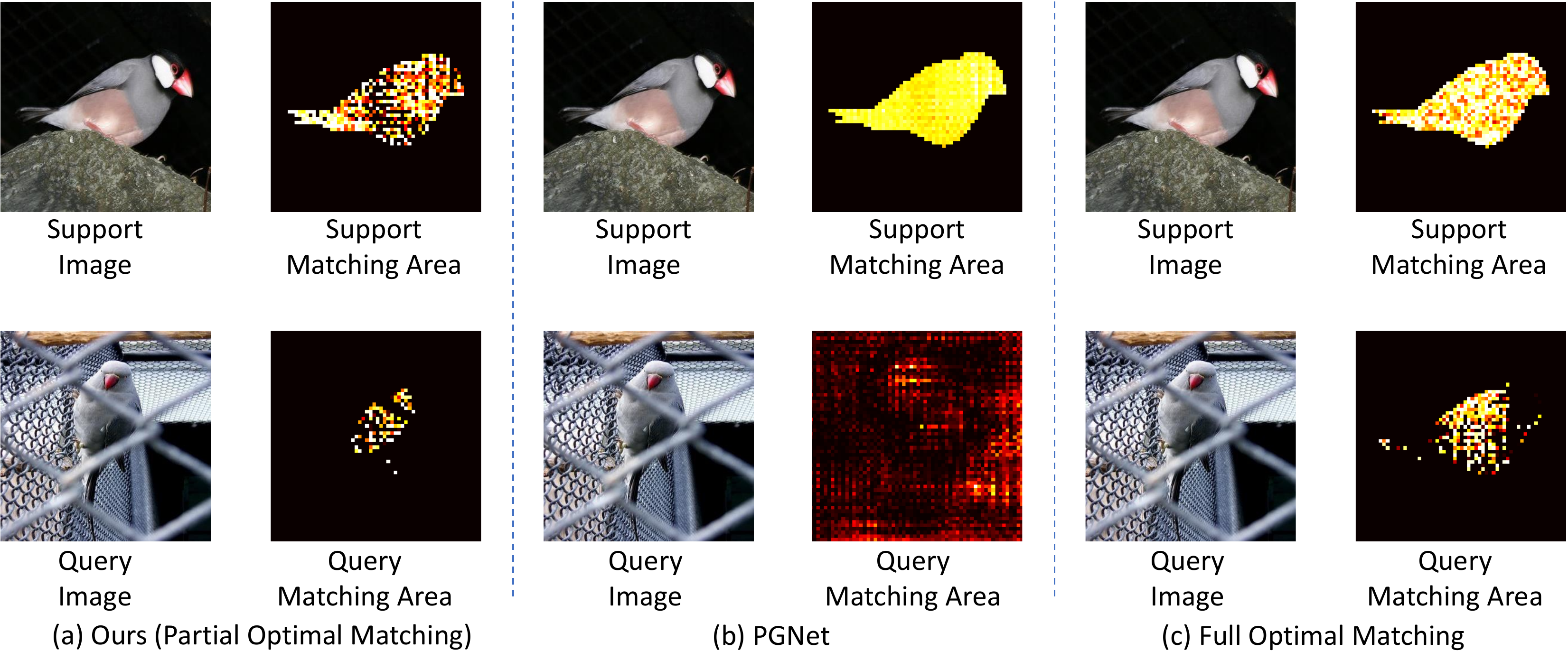}
    \caption{Visualization of the diffident matching methods. The unconstrained many-to-many methods (PGNet~\cite{zhangchi2}) (Figure b) make the network attentive to the most discriminative parts and some non-object regions, making many query features unmatched. In contrast, our proposed constrained many-to-many matching method (partial optimal transport matching) (Figure a) encourages the network to mine more related correspondences accurately. The full optimal transport matching (Figure c) will make some non-object regions to be activated.}
    \label{Fig: pg_ours}
\end{figure*}

\subsection{Message Flow Module}
Previous methods always concentrate on the global information of the objects while overlooking the local feature information. In this section, inspired by SuperGlue~\cite{sarlin2020superglue}, we develop a message flow module to enhance the local feature representations by propagating the messages within one image along inner-flow and between cross images along cross-flow. 

We first employ a shared-parameter CNN to encode the images and model the feature maps as a graph. Each node represents a local feature, and the edges are associated with the similarity between nodes. 
Every initial local feature first aggregates the contextual information by communicating to the connected nodes within the intra-image. Then the local information will be propagated to other images along the cross-flow.

 \begin{figure*}[]
  \centering
    \includegraphics[width=1\linewidth]{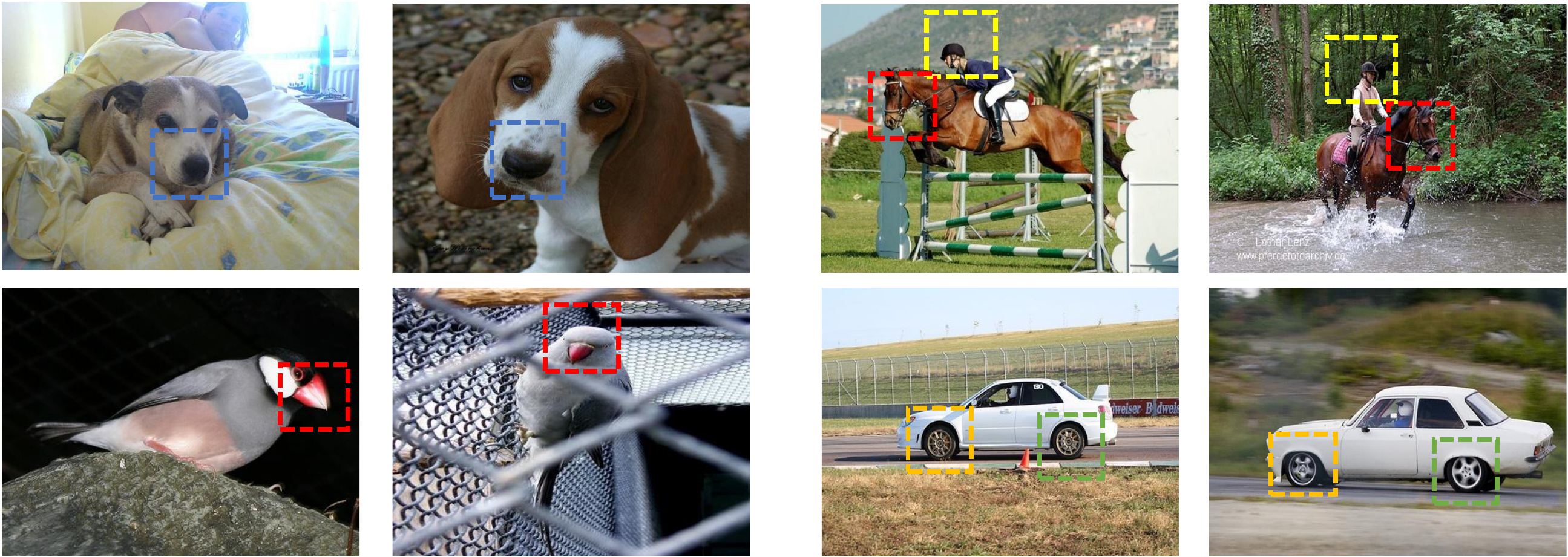}
    \caption{Visualization of the partial optimal matching flows. Given two images (left and right), we plot the best-matched patch of each local region.}
    \label{optimal layer}
\end{figure*}

\textbf{Positional Encoding.}
Besides the object's visual appearance, we argue that the object's position cues can intuitively increase the object's distinctiveness. Inspired by this, we integrate the visual and position information cues as our encoded features. To achieve this, we embed the spatial position information with a learnable absolute spatial positional encoding~\cite{cordonnier2019relationship}. Specifically, for the spatial coordinates (abscissa and ordinate), we independently use sine and cosine functions with different frequencies. Then we fuse the spatial position information and visual appearance features with:

\begin{equation}
    f = f_a + P_{enc}(p_i).
    \label{equation: encoder}
\end{equation}
Where $f_a$ denotes the appearance features, $P_{enc}$ denotes the positional encoding~\cite{cordonnier2019relationship} and $p_i$ denotes the position information. 

\textbf{Inner and Cross-Flow Module.}
Given a pair of graphs whose nodes are the local object features, we consider two kinds of graph message flow propagation~\cite{passing1,passing2,sarlin2020superglue}; the inner-flow within the intra-image and cross-flow between the cross-images. 
We leverage the message passing formulation~\cite{passing1,passing2} to pass on the flow's information.

In particular, we model both support and query image features generated with equation~\ref{equation: encoder} to graphs. For every node $q$ from the graphs, we aggregate all the messages from its connected notes and update information with a residual form:

\begin{equation}
\label{eqn:message-passing}
    f_q' = f_q + MLP(concate(f_q, m{\mathcal{E}\rightarrow q}))),
\end{equation}
here $concate$ denotes the concatenate operation on the channel dimension, and MLP denotes the Multilayer Perceptron (MLP), which implements with $1\times1$ convolution. $m{\mathcal{E}\rightarrow q}$ denotes the aggregation message passing to the node $q$ from its connected nodes, and $f_q$ denotes the feature of the node $q$. For the inner-flow module, the connected nodes are from its neighbor nodes. For cross-flow, we aim to establish a connection between two graphs to propagate the query image information to the support images. We first model the feature maps into graphs with each node corresponding to one feature vector. Each node from the query graph ($n_{qi}$) will be associated with all the support nodes ($n_{s}$). In particular, we calculate the cosine similarity between each query node ($n_{qi}$) and each support node $n_{sj}$, and the edge weight is the similarity between nodes $n_{qi}, n_{sj}$.

\textbf{Aggregation Message Passing.}
We calculate the aggregation message $m{\mathcal{E}\rightarrow q}$ with an attention mechanism.
Inspired by SuperGlue~\cite{sarlin2020superglue}, given a node feature $\mathbf{q}_i$, the keys $\mathbf{k}_j$ and values $\mathbf{v}_j$ from connected nodes, we generate the attention weight as graph edge weight: 
\begin{equation}
    \alpha _{i,j} = SoftMax(\mathbf{q}_{i}^{T}\mathbf{k}_j).
\end{equation}
With the attention weight, we propagate the values from the connected nodes to the node $q_i$ by aggregating the weighted value:
\begin{equation}
    m_{\mathcal E\rightarrow q} =  \sum_{j: (i, j)\in\mathcal{E}} \alpha_{i,j} \mathbf{v}_j.
\end{equation}
We generate the node feature $\mathbf{q}_i$, the keys $\mathbf{k}_j$ and values $\mathbf{v}_j$ with a  $1\times1$ convolution from the query graph and the connected graph. 

We enhance the local feature representations by propagating the message between nodes, benefiting our optimal transport module and the final prediction. 

\subsection{Pre-training on the base class data}
Previous works~\cite{crnet,zhang2019canet,zhangchi2,danet,fss1000,pfenet} always freeze the backbone pretrained from ImageNet~\cite{imagenet}, and train the remaining parameters.  However, a data domain gap exists between the ImageNet to target datasets, which hinders the model from recognizing the objects in the new domain. 
We explore the few-shot segmentation task by learning the pre-training parameters with the base class data. In particular: 1) we first train a model in a fully supervised segmentation way with the training class data by removing all the images containing the validation classes. 2) After that, we replace the backbone parameter with the new parameter generated from step 1 and freeze it. 3) Finally, we fine-tune the remaining parameters the same as the standard few-shot segmentation training step~\cite{zhang2019canet,crnet}.
We utilize a standard cross-entropy loss in the first stage to minimize the distance between the prediction and the ground truth. We optimize the parameters by applying a binary cross-entropy loss in the few-shot segmentation task (third stage) between the query prediction and the target ground truth.
\section{Experiment}

\begin{table}[t]
\centering
\small
\caption{Ablation experiments on the Message Flow Module (MFM), Pre-training on the base classes (Pre-training), Mask-Refine (MR) module~\cite{zhang2019canet}, Prior Mask (PM)~\cite{pfenet} and the Partial Optimize Transport Module (POT). Every module brings performance improvement over the baseline model. The results are reported on PASCAL VOC 2012 dataset with standard mIoU(\%) with the 1-shot setting.
}
\resizebox{0.9\columnwidth}{!}
{
\begin{tabular}{cccccc}
\toprule[1pt]
MFM & Pre-training & MR & PM  & POT & 1-shot \\
\hline
                  &            &            &    &       & 59.1   \\
\checkmark        &            &            &     &      & 60.2   \\
\checkmark      & \checkmark   &            &      &     & 61.0   \\
\checkmark        & \checkmark & \checkmark  &      &    & 61.2   \\
\checkmark       & \checkmark  & \checkmark  & \checkmark &  & 62.0 \\
\checkmark       & \checkmark  & \checkmark  & \checkmark & \checkmark  & 62.5 \\
\bottomrule[1pt]
\end{tabular}

}
\label{table:ablition:different-conponet}
\end{table}

\begin{table}[t]
\centering
\small
\caption{Ablation experiments on the message flow module,  The results reported on PASCAL VOC 2012 dataset with standard mIoU(\%) under the 1-shot setting.
}
\resizebox{0.42\columnwidth}{!}
{
\begin{tabular}{lc}
\toprule[1pt]
Method     & mIoU \\ \hline
Baseline   & 59.1 \\
Inner       & 59.7 \\
Cross      & 59.9 \\
Inner-Cross & 60.2 \\
\bottomrule[1pt]
\end{tabular}

}
\label{table:ablition:attention graph}
\end{table}

\begin{table}[t]
\centering
\small
\caption{Comparison of our Partial Optimal Transport Module (POT), Full Optimal Transport Module (FOT), and the Prior Mask (PM)~\cite{pfenet}. The results are reported on PASCAL VOC 2012 dataset with standard mIoU(\%) under the 1-shot setting.}
\resizebox{0.38\columnwidth}{!}
{
\begin{tabular}{lc}
\toprule[1pt]
Method        & mIoU \\
\hline
Baseline      & 59.1 \\
w/ FOT       & 59.2 \\
w/ POT    & 59.5 \\
w/ PM         & 60.2 \\
w/ FOT-PM    & 59.6 \\
w/ POT-PM & 60.5    \\
\bottomrule[1pt]
\end{tabular}

}
\label{table: ablaition: partial-full-cos}
\end{table}

\begin{table*}[t]
\small
\centering
\caption{Detailed 1-shot and 5-shot results in each split under the mIoU(\%) evaluation metric with PASCAL VOC 2012 dataset. Our model outperforms all previous methods and achieves new state-of-the-art performance.}

\resizebox{0.95\textwidth}{!}{%
\begin{tabular}{lcccccc|ccccc}
\toprule[1pt]

\multirow{2}{*}{Method} & \multirow{2}{*}{Backbone} & \multicolumn{5}{c|}{1-shot} & \multicolumn{5}{c}{5-shot} \\ \cline{3-12} 
& & split-0 & split-1 & split-2 & split-3 & mean & split-0 & split-1 & split-2 & split-3 & mean \\ \hline\hline

FSS-1000~\cite{fss1000} &VGG-16 & -- & -- & -- & -- & -- & 37.4 & 60.9 & 46.6 & 42.2 & 56.8   \\
OSLSM\cite{shaban2017one} &VGG-16 & 33.6 & 55.3 & 40.9 & 33.5 & 40.8 & 35.9 & 58.1 & 42.7 & 39.1 & 43.9 \\

co-FCN\cite{rakelly2018conditional}   &VGG-16      & 36.7   & 50.6   & 44.9   & 32.4   & 41.1  & 37.5   & 50.0   & 44.1   & 33.9   & 41.4 \\
SG-One\cite{zhang2018sg}     &VGG-16    & 40.2  & 58.4   & 48.4   & 38.4   & 46.3 & 41.9   & 58.6   & 48.6   & 39.4   & 47.1 \\

AMP \cite{amp} &ResNet-Wide & 41.9 & 50.2 & 46.7 & 34.7 & 43.4 & 41.8 & 55.5 & 50.3 & 39.9 & 46.9   \\
FWB \cite{fast-weight} &VGG-16 & 47.0 & 59.6 & 52.6 & 48.3 & 51.9 & 50.9 & 62.9 & 56.5 & 50.1 & 55.1  \\
PANet\cite{wang2019panet}   &VGG-16    &42.3 & 58.0   & 51.1   & 41.2   & 48.1 &51.8 & 64.6   & 59.8   & 46.5   & 55.7 \\
CANet~\cite{zhang2019canet} &ResNet-50 & 52.5 &  65.9 &  51.3 &  51.9 &  55.4 &  55.5 &  67.8 &  51.9 &  53.2 &  57.1 \\ 
CRNet~\cite{crnet} &ResNet-50 & 56.8   & 65.8   & 49.4   & 50.6   & 55.7 & 58.7   & 67.9   & 54.2   & 53.5   & 58.8 \\ 

RPMMs~\cite{pmm} & ResNet-50 & 55.1   & 66.9   & 52.6   & 50.7   & 56.3 & 56.2   & 67.3   & 54.5   & 51.0   & 57.3 \\ 

PPNet~\cite{ppnet} & ResNet-101 & 48.5  & 60.1   & \textbf{55.7}   & 46.4   & 52.8 & 58.8   & 68.3   & \textbf{66.7}   & 57.9   & 62.9 \\ 

PFENet~\cite{pfenet} & ResNet-50 & 61.7   & 69.5   & 55.4   & 56.3   & 60.8 & 63.1   & 70.7   & 55.8   & 57.9   & 61.9 \\ 
\hline

Ours  & ResNet-50 & \textbf{65.4}  & \textbf{71.5}   & 55.2   & \textbf{58.1}   & \textbf{62.5} & \textbf{67.0}   & \textbf{71.7}   & 55.8   & \textbf{59.9}   & \textbf{63.6} \\

\bottomrule[1pt]
\end{tabular}
}

\label{Table: Combine-details-voc}
\end{table*}
\subsection{Experiment Setting}
\subsubsection{\textbf{Evaluation Metric}}
Follow previous works~\cite{crnet,shaban2017one}, we adopt the standard mean Intersection-Over-Union(mIoU) as our main evaluation metric. However, ~\cite{Dong2018FewShotSS} also report their results with the mean of foreground IoU and background IoU (FBIoU), which ignore the categories. We report our methods with both evaluation metrics to be a fair comparison.

The evaluation metrics are calculated as follows:

\begin{equation}
    IoU = \frac{Intersection}{Union} = \frac{TP}{TP + FP + FN}, 
    \label{equal:iou}
\end{equation}

\begin{equation}
    mIoU = \frac{1}{n} \sum_{1}^{n}(IoU_{n}),
\end{equation}

\begin{equation}
    FBIoU = \frac{1}{2}(IoU_{fg} + IoU_{bg}).
\end{equation}
where TP denotes true positive, FP denotes false positive; FN denotes false negative, $n$ denotes the classes' number. The standard mIoU is calculated by averaging the IoU of all classes. The $IoU_{fg}$ is calculated with equation~\ref{equal:iou}, which only considers the object foreground and ignores the categories, and $IoU_{bg}$ is calculated in the same way but reversed the foreground and background. FBIoU average the $IoU_{fg}$ and the $IoU_{bg}$. 

\subsubsection{\textbf{Dataset}}
\textbf{PASCAL VOC 2012.}
We validate our methods on the PASCAL VOC 2012 dataset with cross-validation experiments. Follow OSLSM~\cite{shaban2017one}; we split the 20 object categories into 4 folds, three for training and one for testing. For more details about the dataset information and the evaluation metric, please refer to~\cite{shaban2017one}.

\begin{table*}[]
\small
\centering
\caption{Detailed 1-shot and 5-shot results in each split under the mIoU(\%) evaluation metric with the MS COCO dataset. Our model outperforms all previous methods and achieves new state-of-the-art performance.
}

\resizebox{0.95\textwidth}{!}{%
\begin{tabular}{lcccccc|ccccc}
\toprule[1pt]

\multirow{2}{*}{Method} & \multirow{2}{*}{Backbone} & \multicolumn{5}{c|}{1-shot} & \multicolumn{5}{c}{5-shot} \\ \cline{3-12} 
 & & split-0 & split-1 & split-2 & split-3 & mean & split-0 & split-1 & split-2 & split-3 & mean \\ \hline\hline

PANet~\cite{wang2019panet} &VGG-16 & -- & -- & -- & -- & 20.9 & -- & -- & -- & -- & 29.7   \\
FWB\cite{fast-weight}  &VGG-16 & 16.9 & 17.9 & 20.9 & 28.8 & 21.2 & 19.1 & 21.4 & 23.9 & 30.0 & 23.6 \\

RPMMs\cite{pmm}     &ResNet-50    & 29.5   & \textbf{36.8}   & 28.9   & 27.0   & 30.5  & 33.8   & \textbf{41.9}   & 32.9   & 33.3   & 35.5 \\

PFENet~\cite{pfenet} &ResNet-50 & 34.3   & 33.0   & \textbf{32.3}   & 30.1   & 32.4  & 38.5   & 38.6   & \textbf{38.2}   & \textbf{34.3}   & 37.4 \\
\hline

Ours &ResNet-50 & \textbf{48.7}   & 33.3   & 26.8   & \textbf{31.2}   & \textbf{35.0} & \textbf{49.5}   & 35.6   & 31.8  & 33.1   & \textbf{37.5} \\

\bottomrule[1pt]
\end{tabular}
}

\label{Table: coco-soa}
\end{table*}

\begin{table}[t]
\centering
\small
\caption{Comparison with the state-of-the-art methods under the 1-shot and 5-shot setting. Our proposed network outperforms all previous methods and achieves new state-of-the-art performance. The results are reported on PASCAL VOC 2012 dataset with FBIoU(\%) under the 1-shot setting.}
\resizebox{0.89\columnwidth}{!}
{

\begin{tabular}{lccc}
\toprule[1pt]
Method &Backbone & 1-shot (\%) & 5-shot (\%) \\
\hline
OSLM \cite{shaban2017one}    &VGG-16   & 61.3    & 61.5          \\
co-fcn \cite{rakelly2018conditional}  &VGG-16  & 60.9      & 60.2          \\
sg-one  \cite{zhang2018sg}  &VGG-16     & 63.1  & 65.9          \\
AMP \cite{siam2019adaptive}  &ResNet-Wide  & 60.9   & 66.0          \\
PL   \cite{Dong2018FewShotSS}   &VGG-16         & 61.2     & 62.3          \\

CANet \cite{zhang2019canet}  &ResNet-50      & 66.2    & 69.6          \\
CRNet~\cite{crnet}   &ResNet-50   & 66.8    & 71.5     \\
PFENet \cite{pfenet} &ResNet-50 & 73.3    & 73.9         \\
\hline
Ours  &ResNet-50   & \textbf{73.5}    & \textbf{74.1}    \\
\bottomrule[1pt]
\end{tabular}

}

\label{Table: table-combine-shot-bfiou}
\end{table}

\begin{table}[]
\centering
\small
\caption{Comparison with the state-of-the-art methods under the 1-shot and 5-shot setting with FSS-1000 dataset. Our proposed network achieves state-of-the-art performance. The results are reported with standard mIoU(\%) under the 1-shot setting.  
}
\resizebox{0.8\columnwidth}{!}
{

\begin{tabular}{l|c|c|c}
\toprule[1pt]

Method  & Backbone     & 1-shot        & 5-shot        \\
\hline
OSLSM~\cite{shaban2017one}  &VGG-16        & 70.3          & 73.0          \\
co-fcn~\cite{rakelly2018conditional}    &VGG-16     & 71.2          & 74.2          \\
FSS-1000~\cite{fss1000}    &VGG-16    & 73.4        & 80.1          \\
FOMAML~\cite{fomaml}   &Efficient-Net    & 75.1          & 80.6          \\
\hline
Ours & ResNet-50 & \textbf{82.5} & \textbf{83.8}  \\

\bottomrule[1pt]
\end{tabular}
}

\label{table: fss-results}
\end{table}

\begin{table}[t]
\centering
\small
\caption{Comparison with the state-of-the-art methods under the weakly supervised 1-shot setting. Our proposed network achieves state-of-the-art performance. The results are reported on PASCAL VOC 2012 dataset and FSS-1000 dataset with standard mIoU(\%). }
\resizebox{1\columnwidth}{!}
{

\begin{tabular}{l|c|c}
\toprule[1pt]
Method                           & \multicolumn{2}{c}{mIoU(\%)} \\ \hline
                                 & PASCAL     & FSS-1000    \\ \hline

PANet~\cite{wang2019panet} (Bounding box annotations) &  45.1          & -      \\
CANet~\cite{zhang2019canet} (Bounding box annotations) & 52.0       &   -         \\
TeB~\cite{teb} (Bounding box annotations) & -       &   78.2         \\ \hline
Ours (Bounding box annotations) & \textbf{55.4}       & \textbf{80.8}        \\ 
\bottomrule[1pt]
\end{tabular}
}
\label{weakly-1-shot}
\end{table}

\textbf{MS COCO and FSS-1000.}
The major limitation of PASCAL VOC 2012 is that it contains only a few categories that are insufficient to verify the model's capabilities on few-shot segmentation tasks. To evaluate our model more fair and effective, we experiment with our model on more complicated datasets containing more categories and images.

MS COCO 2014~\cite{coco} is a challenging large-scale dataset that contains 80 categories, 82783 training images, and 40504 validation images. Follow the previous work~\cite{pfenet}. We split the 80 object categories into 4 folds, three for training and one for testing.

FSS-1000~\cite{fss1000} introduces a dataset that increases the number of object categories instead of the number of images. In particular, FSS-1000 involves 1000 classes, and each class contains 10 images. Following~\cite{fss1000}, we choose 520 classes for training, 240 classes for validation and 240 classes for testing.

\subsection{Implementation Details}
Our approach exploits multi-level features from the Resnet50 as our feature representations. We adopt dilated convolution to resolve the feature maps not smaller than 1/8 of the input image. We adopt random flipping, random cropping, random rotation, and random scale as data augmentation during training. We utilize SGD as the optimal to train our network with a cross-entropy loss. The initial learning rate is set to 0.0025, and we adopt a poly-policy decay learning rate with a power equal 0.9. 

We train our methods on PASCAL-$5^i$ dataset for 200 epochs and a batch size 4 with an initial learning rate of 0.0025. For the experiment on MS COCO, our models trained for 50 epochs with a batch size of 32 and an initial learning rate of 0.02. For the experiment on FSS-1000, we set the batch size to 4 and an initial learning rate of 0.0025 to train our model for 100 epochs.

\subsection{Ablation study}
In this section, we inspect how each component affects our network and validate the components' performance on the PASCAL VOC 2012 dataset with the 1-shot setting. 
We implement cross-validation experiments in a 1-shot setting and report the results with the standard mean IoU. As shown in Table~\ref{table:ablition:different-conponet}, we establish the network without any of our proposed components as a baseline (The baseline model is directly concatenating the support features to the query features as the final features without any of our proposed components). Each component can bring a significant performance improvement by adding the component one by one. In the beginning, we add the message flow module after the CNN encoder, which can improve 1.1 mIoU over the baseline. Our pre-training on base class data aims to make the network step over the data domain gap; as shown in the table, the pre-training can bring 1.9 mIoU improvement over the baseline. We also adopt the Mask Refinement module proposed in CANet~\cite{zhang2019canet}, which aims to refine the coarse prediction. The mask refinement can only bring 0.2 mIoU improvement; we conjecture that our prediction can already cover most of the discriminative object regions, so the mask-refinement module may not determine more object regions. Following~\cite{pfenet}, we generate a prior mask as a rough foreground probability map to guide the final segmentation, which brings a 0.8 mIoU improvement. Our optimal transport module can identify the most confident object regions. As shown in Table~\ref{table:ablition:different-conponet}, our partial transport optimal module can bring 0.5 mIoU improvement. Our network can improve 3.4 mIoU over the baseline combined with other components. In addition, we also inspect how the Message Flow Module (MFM), and the Partial Optimize Transport module (POT) affect our network on the PASCAL VOC 2012 dataset with 5-shot setting in Table~\ref{table:ablition:different-conponet-5shot}. 

\begin{table}[t]
\centering
\small
\caption{Ablation experiments on the Message Flow Module (MFM), and the Partial Optimize Transport Module (POT). Every module brings performance improvement over the baseline model. The results are reported on PASCAL VOC 2012 dataset with standard mIoU(\%) under the 5-shot setting.
}
\resizebox{0.39\columnwidth}{!}
{
\begin{tabular}{ccc}
\toprule[1pt]
MFM & POT & 5-shot \\ 
\hline

                & \checkmark  & 61.7   \\
\checkmark       &              & 62.1 \\
\checkmark       & \checkmark  & 63.6 \\
\bottomrule[1pt]
\end{tabular}

}
\label{table:ablition:different-conponet-5shot}
\end{table}

\textbf{Inner Flow vs. Cross Flow.}
Table~\ref{table:ablition:attention graph} analysis our model message flow module. Both message flow will bring improvement over the baseline. Combining inner and cross-flow modules will further improve the performance.

\textbf{Iterative message passing vs. stacked message passing.}
Lu \textit{et al.}~\cite{wang1} proposes an attentive graph neural network (AGNN) to iterative fuse the information of different images over the data graphs to extract the common objects from the semantically related images. In particular, the AGNN leverages the ConvGRU to update the node state, where the parameters are shared in each iteration. We also investigate the message passing in a stacked way, which does not share parameters between the different message-flow blocks. 
As shown in Table~\ref{Table:abalation-multi-matching}, we validate both the stacked and the iterative message passing. The performance of stacked message passing decreases when we increase the number of stacked blocks. The stacked message-flow module achieves the best performance with one-step propagation. When we use the message-flow module with multiple blocks, a deeper network is required, which extracts more semantic information, resulting in the message-flow module focusing on the most discriminative parts, leading to poor performance. 
The best performance of iterative message passing is achieved with 5-step propagation. This conclusion is consistent with AGNN.
\begin{table}[]
\centering
\small

\caption{Comparison of the different message flow times. The results are reported on PASCAL VOC 2012 dataset with standard mIoU(\%) under the 1-shot setting.}
\resizebox{0.5\columnwidth}{!}{
\begin{tabular}{l|c|c}
\toprule[1pt]
Number & Stacked & Iterative \\ \hline
One & 62.5 & 62.5\\
Three   & 62.2 & 62.7\\
Five   & 62.1 & 62.8\\ 
Seven   & 61.6 & 62.8\\ 
Nine   & 61.5 & 62.7\\ 
\bottomrule[1pt]
\end{tabular}

\label{Table:abalation-multi-matching}
}
\end{table}

\textbf{Partial Optimal Matching vs.  Full Optimal Matching vs. Other Matching.}
We experiment with a variant by replacing the partial optimal transport with the full optimal transport module to solve the transport problem. As shown in Table~\ref{table: ablaition: partial-full-cos}, both partial and full optimal transport modules can bring improvement over the baseline. 
To investigate how the matching correspondences affect when combined with the prior mask~\cite{pfenet}, we add our optimal matching correspondences after the prior mask to guide our network. The partial optimal correspondences can improve the performances, but the full optimal correspondences drop the performances. We conjecture that the prior mask roughly locates most of the foreground, and our partial matching correspondences locate the confident foreground areas. As shown in Figure~\ref{Fig: pg_ours} (Figure c), the full transport optimal matching may match the wrong object regions.
We also validate the previous matching methods, such as PGNet~\cite{zhangchi2}. The results in Table~\ref{Table:abalation-pg-ours} show that our proposed POT matching achieves better performance. We also visualize the different matching methods in Fig.~\ref{optimal layer} and Fig.~\ref{Fig: pg_ours}

\begin{table}[]
\centering
\small

\caption{Comparison of the different matching methods as our matching features, we replace our proposed POT with the attention mechanism matching proposed by PGNet~\cite{zhangchi2}. The results are reported on PASCAL VOC 2012 dataset with standard mIoU(\%) under the 1-shot setting.}
\resizebox{0.35\columnwidth}{!}{
\begin{tabular}{l|c}
\toprule[1pt]
Method & mIoU \\  \hline
w PGNet & 61.7 \\
w POT   & 62.5 \\
\bottomrule[1pt]
\end{tabular}

\label{Table:abalation-pg-ours}
}
\end{table}

\textbf{The effectiveness of the positional encoding.}
As shown in Table~\ref{Table:abalation-position}, the position encoding brings  an improvement of 0.4 mIoU on PASCAL VOC 2012 dataset with standard mIoU(\%) under the 1-shot setting.
\begin{table}[]
\centering
\small
\caption{Ablation study of positional encoding. PE denotes positional encoding operation. The results are reported on PASCAL VOC 2012 dataset with standard mIoU(\%) under the 1-shot setting.}
\resizebox{0.35\columnwidth}{!}{
\begin{tabular}{l|c}
\toprule[1pt]
Method & mIoU \\  \hline
w/o PE & 62.1 \\
w PE   & 62.5  \\
\bottomrule[1pt]
\end{tabular}

\label{Table:abalation-position}
}
\end{table}

\subsection{Comparison with other state-of-the-art methods}
In this section, we compare our network with other state-of-the-art methods on PASCAL VOC 2012 dataset, MS COCO dataset, and FSS-1000 dataset.

\textbf{PASCAL VOC 2012.} Table~\ref{Table: Combine-details-voc} and Table~\ref{Table: table-combine-shot-bfiou} shows the performance of different methods in both 1-shot and 5-shot setting. We use FBIoU and standard mIoU as the evaluation metric to report our results. The difference between FBIoU and mIoU is that the FBIoU ignores the object category and calculates both foreground and background Intersection-over-Union. We achieve the new state-of-the-art under both 1-shot and 5-shot settings with both FBIoU and mIoU evaluation metrics. 

\textbf{MS COCO.}
As shown in Table~\ref{Table: coco-soa}, we cross-evaluate our methods on the dataset MS COCO~\cite{coco} under both 1- shot and 5-shot setting. Our methods outperform all the previous state-of-the-art methods. 

\begin{figure*}[]
  \centering
    \includegraphics[width=0.9\linewidth]{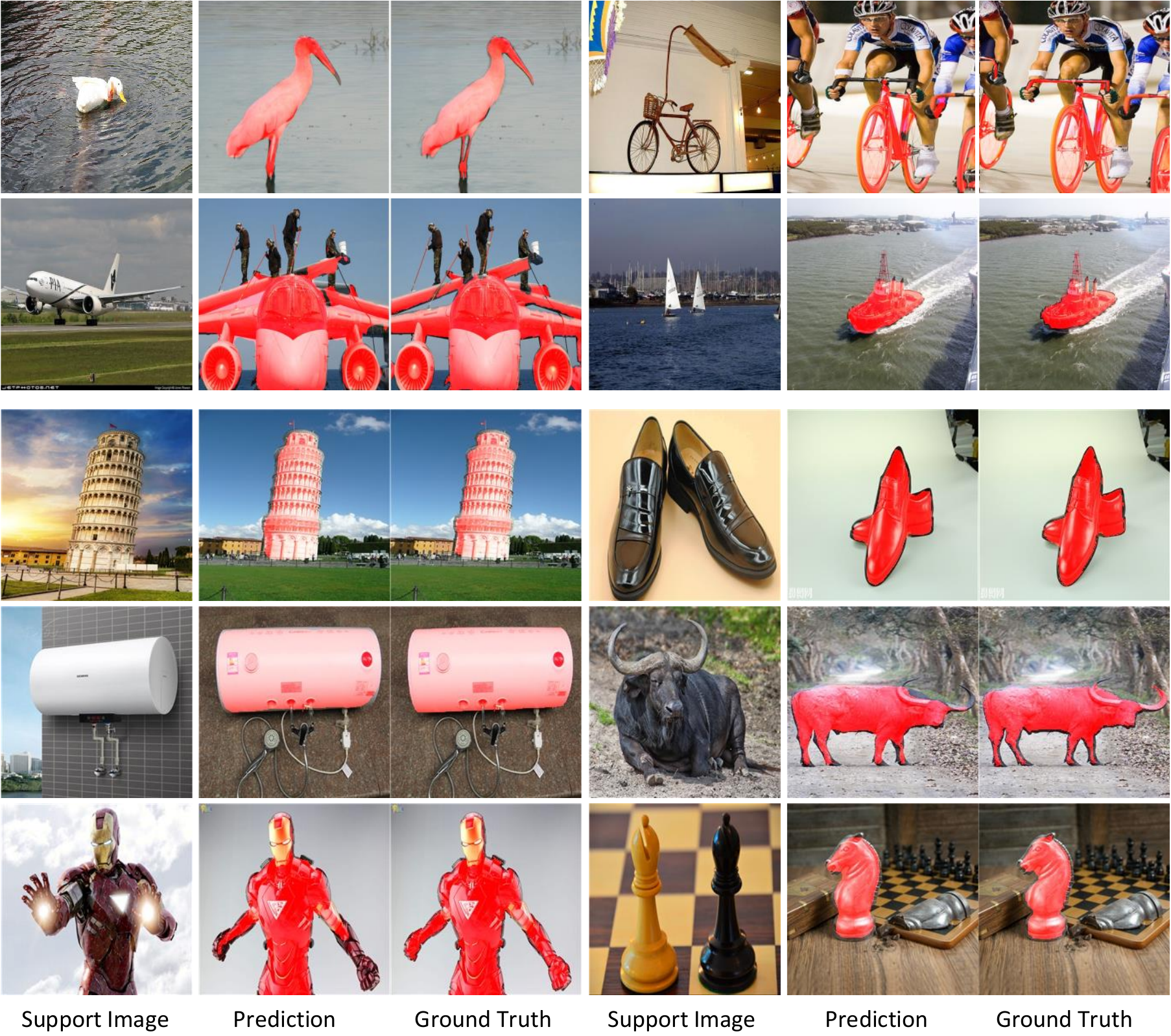}
    \caption{Our qualitative examples on the PASCAL VOC and FSS-1000 dataset. The first two rows are the results from PASCAL VOC, and the last three rows are the results from FSS-1000. The qualitative examples are obtained under the 1-shot setting.}
    \label{combine-qr}
\end{figure*}

\textbf{FSS-1000.}
The comparison results with FSS-1000 dataset~\cite{fss1000} under both 1-shot and 5-shot setting are shown in Table~\ref{table: fss-results}. Our methods outperform all previous methods and achieve new state-of-the-art performance. In addition, we also show the qualitative examples for FSS-1000 and PASCAL VOC2012 in Fig.~\ref{combine-qr}.

\subsection{Weakly supervised few shot segmentation}
Following CANet~\cite{zhang2019canet} and TeB~\cite{teb}, we further evaluate our proposed CMNet with a weaker annotation such as bounding boxes. 
We generate the bounding box annotations from PASCAL VOC 2012, SDS~\cite{hariharan2014simultaneous}, and FSS-1000~\cite{fss1000} dataset by locating every object. During testing, we replace pixel-level annotation with the bounding boxes. 
As is shown in Table~\ref{weakly-1-shot}, our CMNet outperforms all the previous methods and achieves new state-of-the-art in both PASCAL VOC 2012 and FSS-1000 datasets under the weakly supervised few-shot segmentation setting. 
\\

 \begin{figure*}[t]
  \centering
  \vskip +1em
    \includegraphics[width=0.9\linewidth]{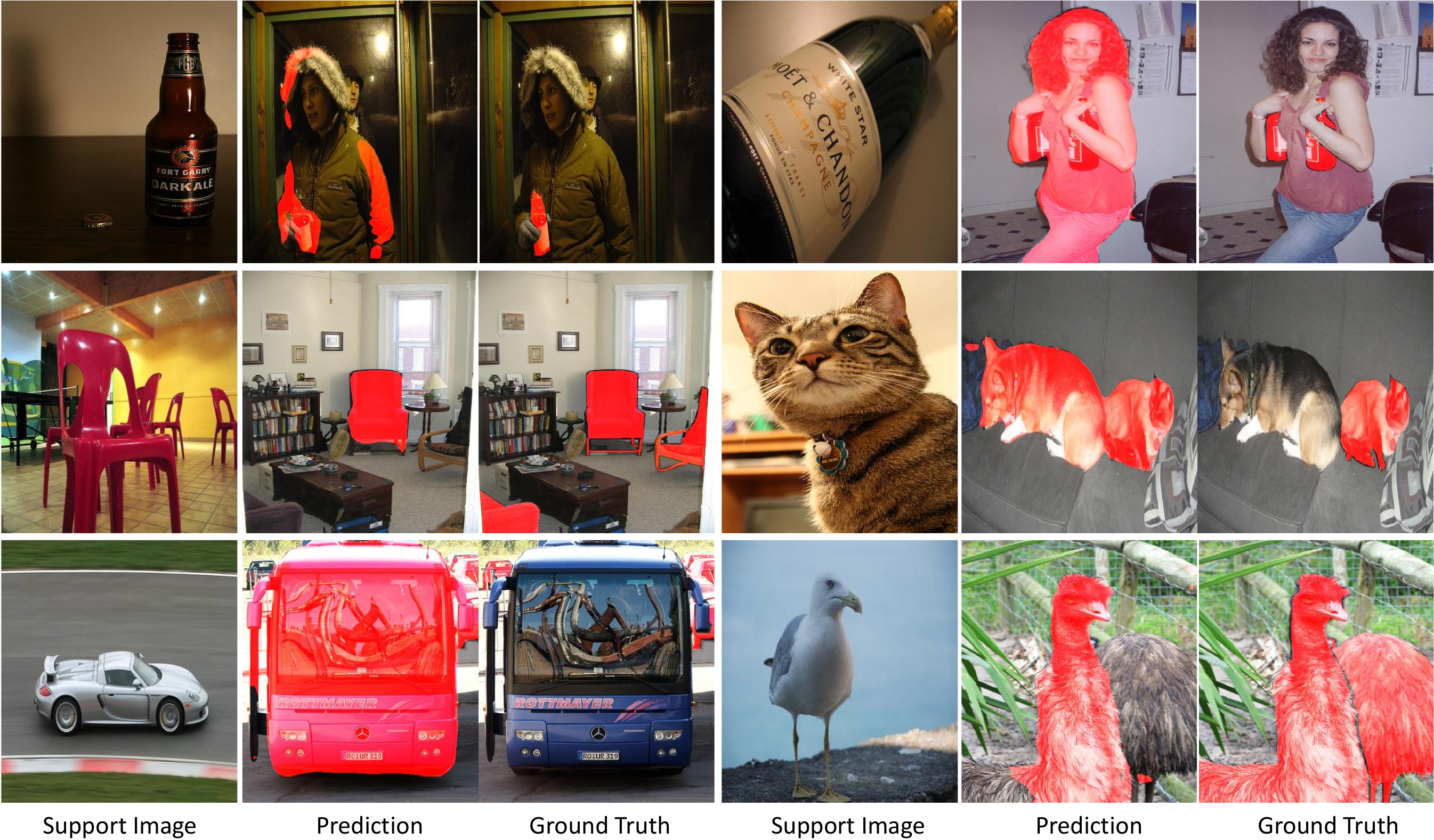}
    \caption{The failure cases on PASCAL VOC 2012 dataset.}
    \label{fail_voc}
\end{figure*}

\subsection{Failure case analysis}
In this section, we analyze the challenging cases that fail our model. As shown in Fig.~\ref{fail_voc}, our model fails to distinguish between dogs and cats, chairs and tables, cars and buses. This is because they have a similar pattern: difficult to distinguish without semantic information. Moreover, our model can not locate the bird bodies because the query image only contains the bird's body information lacking semantic information. 
\section{Conclusion}
This work develops a Correspondences Matching Network for few-shot segmentation. We explore the element-to-element correspondence to the foreground by leveraging a partial optimal transport module and the message flow module to guide our final prediction. 
Experiments on PASCAL VOC 2012, MS COCO, and FSS-1000 dataset validate our contribution.
\section*{Acknowledgements} This research is supported by the National Research Foundation, Singapore under its AI Singapore Programme (AISG Award No: AISG-RP-2018-003), the Ministry of Education, Singapore, under its Academic Research Fund Tier 2 (MOE-T2EP20220-0007) and Tier 1 (RG95/20).

\ifCLASSOPTIONcaptionsoff
  \newpage
\fi

\bibliographystyle{IEEEtran}
\bibliography{egbib}

\end{document}